\documentclass[3p, twocolumn]{elsarticle}
\usepackage{lineno,hyperref}
\usepackage{url}
\usepackage{amsmath}
\usepackage{amssymb}

\usepackage[utf8]{inputenc}
\usepackage{geometry}

\modulolinenumbers[5]

\begin{document}

\begin{frontmatter}

\title{Induction, Popper, and machine learning}

\author{Bruce Nielson}
\ead{brucenielson1@gmail.com}
\author{Daniel C. Elton}
\ead{delton@mgh.harvard.edu}

\begin{abstract}
Francis Bacon popularized the idea that science is based on a process of induction by which repeated observations are, in some unspecified way, generalized to theories based on the assumption that the future resembles the past. This idea was criticized by Hume and others as untenable leading to the famous problem of induction. It wasn’t until the work of Karl Popper that this problem was solved, by demonstrating that induction is not the basis for science and that the development of scientific knowledge is instead based on the same principles as biological evolution. Today, machine learning is also taught as being rooted in induction from big data. Solomonoff induction implemented in an idealized Bayesian agent (Hutter's AIXI) is widely discussed and touted as a framework for understanding AI algorithms, even though real-world attempts to implement something like AIXI suffer immediately encounter fatal problems. In this paper, we contrast frameworks based on induction with Donald T. Campbell’s universal Darwinism. We show that most AI algorithms in use today can be understood as using an evolutionary trial and error process searching over a solution space. In this work we argue that a universal Darwinian framework provides a better foundation for understanding AI systems. Moreover, at a more meta level the process of development of all AI algorithms can be understood under the framework of universal Darwinism. 
\end{abstract}


\end{frontmatter}


\section{Introduction}
Francis Bacon popularized the idea that science was based on a process of induction by which repeated observations are  generalized to theories. This idea was criticized by Hume and others as logically untenable, leading to the famous ``problem of induction'' whereby science was assumed to utilize a process that was logically invalid. It wasn’t until landmark work by Karl Popper that the problem of induction was resolved. By showing how science advances via falsification rather than confirmation Popper showed that induction was not the basis for science. Interestingly, Popper found the growth of scientific knowledge followed the same principles as biological evolution, leading to the field of evolutionary epistemology. Popper claimed to have refuted the idea that induction provides a foundation for knowledge. Years later, many scientists still believe some version of induction (for instance Bayesianism) is the basis for science. Machine learning is also taught as being rooted in induction. Given the success of machine learning, does this mean Popper was wrong that induction is a refuted theory? 
Vague references towards ``inductive learning from data'' are often made, without providing a more explicit understanding of how induction works in machine learning systems. A more concrete formulation of induction, Solomonoff induction, however has been proposed as a basis for AI.\cite{Hutter2005} More generally, the notion that AI systems approximate an idealized Bayesian agent has been quite popular and appears in books such as Nick Bostrom's \textit{Superintelligence}.\cite{Bostrom2014} Solomonoff induction, besides being incomputable and hard to approximate, suffers from several problems,\cite{demski2020embedded,Hutter2009} a few of which we believe are fatal. A full enumeration and study of these problems is beyond the scope of this work (and will the subject of future work), but two particularly fatal issues both arise from the inability to construct an appropriate prior beforehand (the ``grain of truth'' problem and the ``problem of old evidence''). 

What about understanding deep learning as approximating Bayesian statistical modeling? Bayesian neural networks are increasingly discussed and conventional neural networks with dropout have been argued to approximate Bayesian modeling.\cite{Gal2016} We don't think Bayesian statistical modeling provides an adequate framework for understanding deep learning yet alone AI more generally. Gellman and Yao recently have elaborated on how Bayesian statistical modeling suffers from several holes and pitfalls which when encountered practitioners have to grapple with solving through a process of trial and error.\cite{Gelman_2020} After first providing an overview of Universal Darwinism, in this work we argue that a universal Darwinian framework provides a better foundation for understanding AI systems. We discuss how nearly all AI algorithms, with a few exceptions, can be understood as operating within the universal Darwinian frame. Moreover, at a more meta level the process of development of all AI algorithms can be understood using a universal Darwinian framework.

\vspace{-0.5em}
\section{Induction vs universal Darwinism}
\subsection{Francis Bacon and induction}
The concept of induction dates as far back as the 15th century and signifies the idea of specific instances being generalized to universal laws. To use the canonical example, suppose we see a specific swan $S1$ and see that it is colored white, or in other words: 
\begin{equation}
 S1 \rightarrow \mbox{White}
\end{equation}
If we later see a bird that is not white, deductive logic allows us to find that that bird is not $S1$:
\begin{equation}
    \begin{aligned}
        &S1 \rightarrow \mbox{White}\\
        &\neg \mbox{White} \\
        &\therefore \neg S1
    \end{aligned}
\end{equation}

However, deductive logic does not allow us to generalize from a specific statement like this. The fact that $S1$ is white does not allow us to assume that an $S2$ - a different swan - will also be colored white. 

But what if we see hundreds or even thousands of swans and all of them are white? Is there some point at which we can rightly assume that we can now logically reason that all swans are white? In other words, is it valid to reason:
\begin{equation}
    \begin{aligned}
    S1\dots S1000 &\rightarrow \mbox{White}\\
    \therefore \forall_x  Sx  
    \end{aligned}
\end{equation}

The supposed ability to reason from specific statements to universal statements is the method of induction. Francis Bacon popularized the idea that the scientific method was based on this ‘inductive method’ of reasoning from specific statements to universal statements. 

However, Hume pointed out that no matter how many specific statements we observe, we are never justified in reasoning to a universal statement.\cite{Popper1972} And, in fact, the discovery of actual black swans showed that this was the case.  This is because it is logically invalid to ever reason from a specific statement to a universal statement. 

This raised the question: if the inductive method is logically invalid, then how can it be the basis (as Bacon supposed) for scientific discovery? If Bacon was correct, would that then imply that science is ‘unjustified’ and therefore no better than myths and dogmas? These questions soon become known as ‘the problem of induction.’.\cite{Popper1972}

\subsection{Karl Popper's solution}
Philosopher Karl Popper solved the problem of induction by reframing the question entirely. Popper threw out the idea that any sort of ‘justification’ (in the sense of certainty, near-certainty, or even just being probable) was possible. Instead, he believed that the scientific method had nothing to do with induction and was instead based on a Darwinian epistemology (theory of knowledge) where scientists started with some problem they wanted to solve (say, why the perihelion of mercury didn't follow Newton’s laws of physics) and they simply `conjectured' (guessed) possible solutions to the problem (say, Einstein’s special or general relativity) and then subjected both the old and new explanations to tests designed to ‘refute’ either or both theories (say, Arthur Eddington’s expedition to test positions of stars near the sun during an eclipse.) If these critical tests – or more generally any sort of criticism – refutes one of the theories but not the other (say, refuting Newton in favor of Relativity) then it is rational to go with the surviving theory rather than the refuted theory regardless of any need for certainty, justification, or probability. 

Popper argued that this evolutionary ‘survival of the fittest’ theory of knowledge was the actual basis for science and that induction was unnecessary to explain the process and only created unnecessary problems when shoehorned into the scientific method. Popper summarized his view of science as ``conjecture and refutation''.\cite{Popper:LoSD}

The false belief that induction was the basis for science has led to a variety of philosophical mistakes. For example, Baconian induction claims that from many observations (or rather from many specific statements) we generalize to a universal statement or law. But we did not need to observe the perihelion of Mercury thousands of times before we realized something was amiss with Newtonian physics. Often a single observation is sufficient to start the conjecture process so long as the observation is a problem in need of a solution. Therefore, only a special kind of observation – a problem, or in other words an observation at odds with present theories – starts the process to find a new general law. Multiple observations are unnecessary.

But the most important philosophical mistake introduced by Baconian induction was the idea that science ever needed justification or certainty in the first place. Popper pointed out that the mere fact that we can compare two theories via a critical test and demonstrate that one theory/explanation was better than the other was sufficient reason to prefer one theory over the other without ever needing to claim certainty that the theory in question was correct. The mere fact that it is the sole surviving theory currently available to us is reason enough to adopt it. In other words, theories are never confirmed but only falsified, and that's OK. If we can live without confirmation, as Popper argues we can, then we have no need for Baconian induction. 

\subsection{Donald Campbell and universal Darwinism}
In the 1980s philosopher Donald T. Campbell took Popper’s idea of epistemology being rooted in evolutionary processes and found a way to generalize it. Campbell claimed that not only was science based on an evolutionary process of ‘survival of the fittest (idea)’ but in fact all knowledge creation was based on evolutionary epistemology.\cite{Campbell1988} Popper later strongly endorsed this generalization of his own theory.\cite{Popper1988} This generalization was later referred to as ``universal Darwinism''.\cite{Campbell2009} The universal Darwin meta-algorithm can be thought of as a generalization of biological evolution. Typically, biological evolution is understood as having three steps:
\begin{enumerate}
\item Replication of genes.
\item Inheritance of a phenotype from the genes with some random variation among the offspring due to either mutation or sexual crossover.
\item Differential survival of the offspring according to which have a phenotype best at replicating the next generation.
\end{enumerate}

Our proposed version of the universal Darwinism algorithm (influenced heavily by Campbell) makes two  important generalizations. The first is that it does not require variations to be randomized, though they may be. For example, trying out a full sweep of all possible variations non-randomly is acceptable as well. So does trying out only variations that match some heuristic or criteria meant to narrow the search process. Campbell suggested referring to this broader category of possible variations as “blind variation” to distinguish it from the more narrow concept of “random variation".\cite{Campbell1988} However, this term tends to be misleading and has led to confusion in the past.\cite{CZIKO1998} So we favor instead simply calling this process “variation and selection” with no qualifier in front of the word “variation.” 

We propose that any process where variant solutions to a problem are generated -- by any means -- followed by a selection process to narrow down to the best variations, will be considered consistent with our version of “universal Darwinism” regardless of how the variants were generated. 

The second generalization is that we do not require replicators or inheritance from a previous generation. To use an example from Campbell\cite{Campbell1988}, a paramecium that is blocked so that it can’t move forward will try out each possible direction it can. Once it finds a direction that is not blocked, it will retain that direction until again blocked. The ‘variations’ are the different directions tried. The direction that it finds to not be blocked is the ‘selected’ variant. So Campbell considered this example to be an evolutionary process despite there being no ‘inheritance’ from one variant to another. Of course, having inheritance also is acceptable under “universal Darwinism”, it just is no longer considered a requirement as in biological evolution.
The final Universal Darwin algorithm is therefore simply “variation and selection” though to put a finer point on it we’ll summarize this meta-algorithm as:

\begin{enumerate}
\item Start with a problem
\item Conjecture a solution(s) to the problem (i.e. create variants)
\item Measure how well the proposed solution(s) solve the problem
\item Over time, retain the better solution(s)
\item Repeat from step 2 until the problem is sufficiently solved or an optima is reached
\end{enumerate}

The “over time” in step 4 simply means that the algorithm doesn’t need to always pick the best variation. It must merely do so over some unspecified period of time. This allows algorithms like simulated annealing to be considered valid kinds of universal Darwinism.

Many seemingly different algorithms will qualify as universal Darwinism. Obviously genetic algorithms will qualify, but so will all search algorithms. Moreover, Popper’s scientific epistemology and biological Darwinian natural selection are now just a subset of this more general algorithm. This meta-algorithm thereby generalizes every type of evolutionary algorithm that we currently know of. That is the sense in which it is to be considered “universal.” 

Universal Darwinism works for the very simple reason that if you compare two (or more) variants as possible solutions to a problem and keep the better solutions – while discarding the worse ones – ultimately you will select and retain the best variants, at least until a maxima is reached. Whether or not it is a global or local maxima depends on the specifics of the problem and the chosen algorithm. For our purposes of this paper, we will refer to this process of improvement of variants as \emph{knowledge-creation}. 

\section{Artificial Intelligence and universal Darwinism}
\subsection{All search algorithms utilize universal Darwinism}

Unlike Popper’s epistemology, which was specific to how we improve scientific explanations, this final algorithm also applies to other kinds of knowledge creation including searching for optimal solutions to a problem or finding useful heuristics. Russell and Norvig suggested that artificial intelligence be measured in terms of how well it finds an optimal or (if optimality is intractable) near optimal solution to a problem.\cite{Russel2010} As such universal Darwinism seems ideal for the kinds of computational problems artificial intelligence is trying to solve so it should be of interest to the fields of artificial intelligence and machine learning. It is of added interest that the universal Darwinism meta-algorithm unifies these types of AI algorithms with biological evolution as well as human thought and culture (via Popper’s epistemology.) So both narrow AI and AGI are types of universal Darwinism.

A survey of Russel \& Norvig,\cite{Russel2010} the most popular introductory text on AI, finds that many existing AI algorithms already utilize universal Darwinism as the basis for how they work. For example, all of the following popular AI algorithms are really evolutionary algorithms of variation and selection and thus fall under the Universal Darwin algorithm:

\begin{enumerate}
\item Problem Solving by Searching \\
{\small Example: A-Star works by trying out every possible variant (often guided by an admissible heuristic) until the shortest path is found.}
\item Non-Optimal Search Algorithms \\
{\small Example: hill climbing, gradient descent, simulated annealing, and genetic algorithms all utilize variants and selection of the best variants.}
\item Adversarial Search \\
{\small Example: Minimax algorithms search every possible move utilizing a heuristic such as a board evaluation algorithm as a proxy for the best board position.}
\item Constraint Satisfaction Problems \\
{\small Example: Many CSP algorithms try out each possible combination to find an acceptable solution.}
\item Logic and Planning Problems \\
{\small Example: The DPLL algorithm does recursive depth-first search enumerating possible models.}
\end{enumerate}

Likely, most people working within the AI field have never thought of search algorithms as being a kind of evolutionary algorithm of variation and selection. It is common to only think of genetic algorithms as being evolutionary algorithms due to those being the only kind that match biological evolution as described in section C above. Yet all search algorithms count as forms of the more generalized universal Darwin algorithm. Considering the prominence of search algorithms within the field of AI, this underscores the possible value of rethinking AI from within the paradigm of universal Darwinism. 

This also raises a fascinating question. Is it possible that universal Darwinism powers every kind of knowledge creation? Or is universal Darwinism simply one of several possible ways to create knowledge? 

Campbell predicted as far back as 1960~\cite{Campbell1988} that the universal Darwin algorithm was the sole source of knowledge creation and that any time we had a discovery or expansion to our knowledge we’d find that variation and selection was requisite.\cite{Campbell1988} So if we could find, within the AI field, any counterexamples, that would be of interest to the field of universal Darwinism as it would challenge the view that universal Darwinism is a sole source of knowledge-creation.
	
\subsection{Machine learning and induction}

While search is prominent in existing AI algorithms, and by extension so is universal Darwinism, it is less clear if this is also the case for AI’s most popular branch: machine learning. 

Machine learning grew out of Statistical Learning techniques such as Linear and Logistic Regression. Statistics is often considered an “inductive” process. Because of this shared history, machine learning is also usually framed in terms of induction. For example, Mitchell frames machine learning in terms of “The Inductive Learning Hypothesis”:
\begin{quote}
``Any hypothesis found to approximate the target function well over a sufficiently large set of training examples will also approximate the target function well over other unobserved examples''.\cite{Mitchell1997} 
\end{quote}

How can machine learning’s inductive roots be squared with Popper’s refutation of Induction?
	
\subsection{Baconian induction vs statistical induction}
 
Because Baconian Induction was (wrongly) intended to describe how science worked, some of the language of science has slipped into statistical induction. For example, in both statistics and machine learning we often refer to a proposed model as a ‘hypothesis’ as if it is a scientific theory. But statistical/machine learning models rarely take the form of an explanatory theory and are generally simple predictive heuristics.\cite{Elton2021} 

To make this distinction more concrete, whereas Baconian induction endeavours to reason from observing several white swans to a universal statement “All swans are white”, statistical induction sets a lower bar by using a random sample to predict how common white swans are (a shift from universal statements about the world to multiple statements with credences assigned to each one). 

While statistical induction has some utility, it also has its own set of problems. For example, if you lived in Europe during the 16th century when it was believed that all swans must be white (and thus “black swan” had come to refer to something being “impossible”) even a seemingly valid random frequentist sample would have found 100\% of swans to be white because black swans were only available in the yet to be discovered Australian continent. So we see that even a statistical inductive model only makes good predictions if you first have a correct prior theory about what variables to factor over, such as in this case location. Statistical models are inherently parochial, they do not have ``reach'' beyond the domain where the sampling is taking place.\cite{Deutsch2011-DEUTBO} Models which do not reach beyond what is known cannot stake out new claims (conjectures) and thus do not expose themselves to refutation. Statistical induction thus cannot replace the conjecture and refutation process of knowledge creation and rather is just a tool which is sometimes useful when building theories that do have reach. 

\begin{table}[]
    \centering
    \small
    \begin{tabular}{|p{3.5cm}|p{3.5cm}|}
    \hline
    \multicolumn{2}{|c|}{\textbf{Table 1: Two kinds of induction }}\\   \hline 
 \textbf{Baconian Induction} & \textbf{Statistical Induction} \\ \hline
Logically reasons from specific statements to universal statements    
  & Based on frequentism \\ \hline
Obtains certainty, or at least 'justification' for knowledge
  & Offers no certainty nor justification \\ \hline
 Is the basis for scientific explanatory hypotheses & 
 Is not the basis for scientific explanatory hypotheses. Instead it creates useful heuristics.\\ \hline
 Doesn't exist & Exists \\
 \hline
    \end{tabular}
    \label{tab:kindsofinduction}
\end{table}

\subsection{Machine learning and universal Darwinism}
Machine learning is arguably the most important form of AI and, particularly with deep learning, the most successful. Often these functions are of a nature that no human being knows how to program them directly. For example, we don’t really know how to build a really good facial recognition algorithm using traditional programming techniques, but deep learning can create one for us that has a high degree of accuracy. It is unnecessary for us to know exactly how deep learning creates such algorithms to work effectively enough for many purposes. It is currently an active area of research to understand why deep learning works so well.\cite{ChoromanskaLeCun2015}

So is deep learning really based on statistical induction, as is widely assumed, or is it rooted in universal Darwinism? Or is it a mixture of both? While it may well be a mixture of both, we argue that the evolutionary aspects of a typical gradient descent algorithm are key to understanding why deep learning creates knowledge. Deep learning typically involves algorithms similar this:

\begin{enumerate}
    \item Outer Loop:
        \begin{enumerate}
        \item Initialize network weights randomly
        \item \emph{Try a set of hyper parameters}
        \item Inner Loop
            \begin{enumerate}
            \item Measure loss function for weights
            \item Calculate slope at current weights
            \item Use slope to \emph{try to move to a better set of weights}
            \item Go to step i until improvements on loss function stop for some period of time.
            \end{enumerate} 
        \item Go back to step b
        \end{enumerate} 
\end{enumerate}

Most people may not think of deep learning as an evolutionary algorithm but a careful look at the above algorithm (where emphasized) reveals deep learning is actually two nested evolutionary algorithms of trying variant solutions. This corresponds nicely with Campbell’s “nested hierarchy of selective retention processes”\cite{Campbell1988} where one evolutionary process can drive another evolutionary process. For instance, at one level the evolution of ideas may happen within a given mind, while at a higher level the evolution of minds is happening via biological evolution. Noble and Noble suggested that this hierarchy of evolution is the basis for why evolution often seems ``purposeful'' due to one level of evolution driving another level towards some goal or purpose.\cite{Noble2017} An example of this hierarchy of evolution is how the immune system can purposefully drive hypermutation of genes to find the correct antibodies for a particular invader.\cite{Noble2017} 

Increasingly, the training of deep learning models is being understood within the framework of ``search''. The lottery ticket hypothesis, for instance, suggests that much of what stochastic gradient descent is doing is finding the best subnetwork amongst the random weights of the initial network.\cite{Frankle2019} Recently Ramanujan et al. have shown that large networks with random weights have highly preferment subnetworks within them, or in other words, that good performance can be achieved merely by identifying a subnetwork rather than changing any weights.\cite{Ramanujan2020}

\subsection{Is the development of AI occurring in accordance with universal Darwinism?}
It is instructive that deep learning seems like an inductive process but actually also utilizes an evolutionary algorithm. Is it possible that all AI systems are subsumed by the universal Darwinism framework? 

Most machine learning algorithms that involve optimization can be subsumed into the framework. As a random example, an ID3 decision tree tries out each feature and measures the entropy information gain for each feature and then selects the best one. However it would seem that not every machine learning algorithm relies on universal Darwinism. Consider the na\"{i}ve Bayes classifier as a counterexample:

\begin{equation}
    \begin{aligned}
        P(Cause, Effect_{1},...Effect_{n}) = 
        \\P(Cause)\prod _{i}P(Effect_{i}|Cause))
    \end{aligned}
\end{equation}

This formula seems to be purely ‘statistical induction’, a fixed process by which probability distributions are updated as new data is fed into the system. It is important to emphasize, however, that any successful application of Bayesian updating must be nested within an evolutionary process.\cite{Gelman2012}

The selection of the right prior is recognized as a problem which is fatal to the idea that Bayesianism can operate successfully within a closed framework. Rather, a proper prior must be fed in and that prior must be informed by knowledge. If the prior is too narrow, Bayesian updating is not guaranteed to converge and if it does converge may yield non-sensical answer. Priors which are too broad, on the other hand, can lead to Goodman’s new riddle of induction. Priors that are too weak (flat) can lead to overconfidence.\cite{Gelman2020} A full discussion of these points is outside the scope of this paper but will be the focus of a soon to be published work.\cite{Elton2021PoC} 

The point is that the successful construction of any useful Bayesian model itself involves a trial and error process – what Andrew Gelman calls “the data analysis cycle”.\cite{Gelman2012} We also note that in passing that na\"{i}ve Bayes strongly under performs on most machine learning tasks. For example, using a na\"{i}ve Bayes classifier on MNIST has an error rate approach 20\% whereas a recent deep learning model obtained an error rate of 0.21\%.\cite{MNISTboard}

Perhaps even more interesting are algorithms that can optionally use universal Darwinism. The best example of this is linear regression. Linear regression typically is done using gradient descent, which we already demonstrated is an evolutionary algorithm. But linear regression can also be performed using the normal equations which do not use an evolutionary algorithm. The reason why we tend to prefer using gradient descent for linear regression is because the normal equations quickly become intractable. This suggests that at least one advantage to utilizing universal Darwinism is tractability. By trying out variants, perhaps guided by a good heuristic search such as gradient descent, we can find good approximate solutions to a problem that would otherwise be intractable to solve for if the entire set of possible solutions had to be searched. This is the whole basis for non-optimal search algorithms like hill climbing or simulated annealing but also for gradient descent. These counterexamples suggest that Campbell may be incorrect that evolutionary algorithms (universal Darwinism) are the only means by which knowledge can be created. 

\section{Conclusion}
We have seen that most artificial intelligence algorithms in use today can be understood within a universal Darwinian framework involving a process of variation and selection that searches for a local or global maxima solution to a problem. Though there are some exceptions to this, such as na\"{i}ve Bayes, the algorithms that do not utilize evolutionary algorithms currently underperform algorithms that do utilize an evolutionary approach. There is as of yet no universally agreed upon theoretical foundation for understanding how AI systems are trained/created but a popular approach is to understand things in terms as some approximation idealized Bayesian induction. In contrast to induction based foundations, which are riddled with problems, we believe that Universal Darwinism provides a stronger and more general foundation for understanding the emergence of intelligent algorithms.  

\subsection*{Funding \& disclaimer}
No funding sources were used in the creation of this work. The authors wrote this article in their personal capacity. 
\vspace{-1em}
%
\bibliography{bibliography.bib}

\end{document}